# Expectation Propagation for Approximate Bayesian Inference

**Thomas P. Minka**
Statistics Dept.
Carnegie Mellon University
Pittsburgh, PA 15213

## Abstract

This paper presents a new deterministic approximation technique in Bayesian networks. This method, "Expectation Propagation," unifies two previous techniques: assumed-density filtering, an extension of the Kalman filter, and loopy belief propagation, an extension of belief propagation in Bayesian networks. Loopy belief propagation, because it propagates exact belief states, is useful for a limited class of belief networks, such as those which are purely discrete. Expectation Propagation approximates the belief states by only retaining expectations, such as mean and variance, and iterates until these expectations are consistent throughout the network. This makes it applicable to hybrid networks with discrete and continuous nodes. Experiments with Gaussian mixture models show Expectation Propagation to be convincingly better than methods with similar computational cost: Laplace's method, variational Bayes, and Monte Carlo. Expectation Propagation also provides an efficient algorithm for training Bayes point machine classifiers.

## 1 INTRODUCTION

Bayesian inference is often hampered by large computational expense. Fast and accurate approximation methods are therefore very important and can have great impact. This paper presents a new deterministic algorithm, Expectation Propagation, which achieves higher accuracy than existing approximation algorithms with similar computational cost.

Expectation Propagation is an extension of *assumed-density filtering* (ADF), a one-pass, sequential method for computing an approximate posterior distribution. In ADF, observations are processed one by one, updating the posterior distribution which is then approximated before processing the next observation. For example, we might replace the exact one-step posterior with a Gaussian having the same mean and same variance (Maybeck, 1982; Opper & Winther, 1999). Or we might replace a posterior over many variables with one that renders the variables independent (Boyen & Koller, 1998). The weakness of ADF stems from its sequential nature: information that is discarded early on may turn out to be important later. ADF is also sensitive to observation ordering, which is undesirable in a batch context.

Expectation Propagation (EP) extends ADF to incorporate *iterative refinement* of the approximations, by making additional passes through the network. The information from later observations refines the choices made earlier, so that the most important information is retained. Iterative refinement has previously been used in conjunction with sampling (Koller et al., 1999) and extended Kalman filtering (Shachter, 1990). Expectation Propagation is faster than sampling and more general than extended Kalman filtering. It is more expensive than ADF by only a constant factor—the number of refinement passes (typically 4 or 5). EP applies to all statistical models to which ADF can be applied and, as shown in section 3.2, is significantly more accurate.

In belief networks with loops it is known that approximate marginal distributions can be obtained by iterating the belief propagation recursions, a process known as loopy belief propagation (Frey & MacKay, 1997; Murphy et al., 1999). In section 4, this turns out to be a special case of Expectation Propagation, where the approximation is a completely disconnected network. Expectation Propagation is more general than belief propagation in two ways: (1) like variational methods, it can use approximations which are not completely disconnected, and (2) it can impose useful constraints on functional form, such as multivariate Gaussian.

## 2 ASSUMED-DENSITY FILTERING

This section reviews the idea of assumed-density filtering (ADF), to lay groundwork for Expectation Propagation. Assumed-density filtering is a general technique for computing approximate posteriors in Bayesian networks and other statistical models. ADF has been independently proposed in the statistics (Lauritzen, 1992), artificial intelligence (Boyen & Koller, 1998; Opper & Winther, 1999), and control (Maybeck, 1982) literatures. "Assumed-density filtering" is the name used in control; other names include "online Bayesian learning," "moment matching," and "weak marginalization." ADF applies when we have postulated a joint distribution $p(D, \mathbf{x})$ where $D$ has been observed and $\mathbf{x}$ is hidden. We would like to know the posterior over $\mathbf{x}$, $p(\mathbf{x}|D)$, as well as the probability of the observed data (or evidence for the model), $p(D)$. The former is useful for estimation while the latter is useful for model selection.

For example, suppose we have observations from a Gaussian distribution embedded in a sea of unrelated clutter, so that the observation density is a mixture of two Gaussians:

$$p(\mathbf{y}|\mathbf{x}) = (1-w)\mathcal{N}(\mathbf{y};\mathbf{x},\mathbf{I}) + w\mathcal{N}(\mathbf{y};0,10\mathbf{I})$$
$$\mathcal{N}(\mathbf{y};\mathbf{m},\mathbf{V}) = \frac{\exp(-\frac{1}{2}(\mathbf{y}-\mathbf{m})^T\mathbf{V}^{-1}(\mathbf{y}-\mathbf{m}))}{|2\pi\mathbf{V}|^{1/2}}$$

The first component contains the parameter of interest, while the other component describes clutter. $w$ is the known ratio of clutter. Let the $d$-dimensional vector $\mathbf{x}$ have a Gaussian prior distribution:

$$p(\mathbf{x}) \sim \mathcal{N}(0, 100\mathbf{I}_d) \qquad (1)$$

The joint distribution of $\mathbf{x}$ and $n$ independent observations $D = \{\mathbf{y}_1, ..., \mathbf{y}_n\}$ is therefore:

$$p(D, \mathbf{x}) = p(\mathbf{x}) \prod_i p(\mathbf{y}_i|\mathbf{x}) \qquad (2)$$

The Bayesian network for this problem is simply $\mathbf{x}$ pointing to the $\mathbf{y}_i$. But we cannot use belief propagation because the belief state for $\mathbf{x}$ is a mixture of $2^n$ Gaussians. To apply ADF, we write the joint distribution $p(D, \mathbf{x})$ as a product of terms: $p(D, \mathbf{x}) = \prod_i t_i(\mathbf{x})$ where $t_0(\mathbf{x}) = p(\mathbf{x})$ and $t_i(\mathbf{x}) = p(\mathbf{y}_i|\mathbf{x})$. Next we choose an approximating family. In the clutter problem, a spherical Gaussian distribution is reasonable:

$$q(\mathbf{x}) \sim \mathcal{N}(\mathbf{m}_x, v_x\mathbf{I}_d) \qquad (3)$$

Finally, we sequence through and incorporate the terms $t_i$ into the approximate posterior. At each step we move from an old $q^{\setminus i}(\mathbf{x})$ to a new $q(\mathbf{x})$. (To reduce notation, we drop the dependence of $q(\mathbf{x})$ on $i$.) Initialize with $q(\mathbf{x}) = 1$. Incorporating the prior term is trivial, with no approximation needed. To incorporate a more complicated term $t_i(\mathbf{x})$, take the exact posterior

$$\hat{p}(\mathbf{x}) = \frac{t_i(\mathbf{x})q^{\setminus i}(\mathbf{x})}{\int_\mathbf{x} t_i(\mathbf{x})q^{\setminus i}(\mathbf{x})d\mathbf{x}} \qquad (4)$$

and minimize the KL-divergence $D(\hat{p}(\mathbf{x})||q(\mathbf{x}))$ subject to the constraint that $q(\mathbf{x})$ is in the approximating family. This is equivalent to a maximum-likelihood problem with data distribution $\hat{p}$. For a spherical Gaussian, the solution is given by matching moments:

$$E_q[\mathbf{x}] = E_{\hat{p}}[\mathbf{x}] \qquad (5)$$
$$E_q[\mathbf{x}^T\mathbf{x}] = E_{\hat{p}}[\mathbf{x}^T\mathbf{x}] \qquad (6)$$

With any exponential family, ADF reduces to propagating expectations. Each step also produces a normalizing factor $Z_i = \int_\mathbf{x} t_i(\mathbf{x})q^{\setminus i}(\mathbf{x})d\mathbf{x}$. The product of these normalizing factors estimates $p(D)$. In the clutter problem, we have

$$Z_i = (1-w)\mathcal{N}(\mathbf{y}_i; \mathbf{m}_x^{\setminus i}, (v_x^{\setminus i}+1)\mathbf{I}) + w\mathcal{N}(\mathbf{y}_i; 0, 10\mathbf{I}) \quad (7)$$

The final ADF algorithm is:

1. Initialize $\mathbf{m}_x = 0$, $v_x = 100$ (the prior). Initialize $s = 1$ (the scale factor).

2. For each data point $\mathbf{y}_i$, update $(\mathbf{m}_x, v_x, s)$ according to

$$s = s^{\setminus i} \times Z_i$$
$$r_i = 1 - \frac{1}{Z_i}w\mathcal{N}(\mathbf{y}_i; 0, 10\mathbf{I})$$
$$\mathbf{m}_x = \mathbf{m}_x^{\setminus i} + v_x^{\setminus i} r_i \frac{\mathbf{y}_i - \mathbf{m}_x^{\setminus i}}{v_x^{\setminus i} + 1}$$
$$v_x = v_x^{\setminus i} - r_i \frac{(v_x^{\setminus i})^2}{v_x^{\setminus i} + 1} + r_i(1-r_i)\frac{(v_x^{\setminus i})^2 \|\mathbf{y}_i - \mathbf{m}_x^{\setminus i}\|^2}{d(v_x^{\setminus i}+1)^2}$$

This algorithm can be understood in an intuitive way: for each data point we compute its probability $r$ of not being clutter, make a soft update to our estimate of $\mathbf{x}$ ($\mathbf{m}_x$), and change our confidence in the estimate ($v_x$). However, it is clear that this algorithm will depend on the order in which data is processed, because the clutter probability depends on the current estimate of **x**.

## 3 EXPECTATION PROPAGATION

This section describes the Expectation Propagation algorithm and demonstrates its use on the clutter problem. Expectation Propagation is based on a novel interpretation of assumed-density filtering. ADF was described as treating each observation term $t_i$ exactly and then approximating the posterior that includes $t_i$. But we can also think of it as first approximating $t_i$ with some $\tilde{t}_i$ and then using an exact posterior with $\tilde{t}_i$. This interpretation is always possible because we can define the approximate term $\tilde{t}_i$ to be the ratio of the new posterior to the old posterior times a constant:

$$\tilde{t}_i(\mathbf{x}) = Z_i \frac{q(\mathbf{x})}{q^{\backslash i}(\mathbf{x})} \quad (8)$$

Multiplying this approximate term by $q^{\backslash i}(\mathbf{x})$ gives $q(\mathbf{x})$, as desired. An important property is that if the approximate posterior is in an exponential family, then the term approximations will be in the same family.

The algorithm of the previous section can thus be interpreted as sequentially computing a Gaussian approximation $\tilde{t}_i(\mathbf{x})$ to every observation term $t_i(\mathbf{x})$, then combining these approximations analytically to get a Gaussian posterior on **x**. Under this perspective, the approximations do not have any required order—the ordering only determined how we *made* the approximations. We are free to go back and refine the approximations, in any order. This gives the general form of Expectation Propagation:

1. Initialize the term approximations $\tilde{t}_i$

2. Compute the posterior for **x** from the product of $\tilde{t}_i$:

$$q(\mathbf{x}) = \frac{\prod_i \tilde{t}_i(\mathbf{x})}{\int \prod_i \tilde{t}_i(\mathbf{x}) d\mathbf{x}} \quad (9)$$

3. Until all $\tilde{t}_i$ converge:

   (a) Choose a $\tilde{t}_i$ to refine
   (b) Remove $\tilde{t}_i$ from the posterior to get an 'old' posterior $q^{\backslash i}(\mathbf{x})$, by dividing and normalizing:

   $$q^{\backslash i}(\mathbf{x}) \propto \frac{q(\mathbf{x})}{\tilde{t}_i(\mathbf{x})} \quad (10)$$

   (c) Combine $q^{\backslash i}(\mathbf{x})$ and $t_i(\mathbf{x})$ and minimize KL-divergence to get a new posterior $q(\mathbf{x})$ with normalizer $Z_i$.
   (d) Update $\tilde{t}_i = Z_i q(\mathbf{x})/q^{\backslash i}(\mathbf{x})$.

4. Use the normalizing constant of $q(\mathbf{x})$ as an approximation to $p(D)$:

$$p(D) \approx \int \prod_i \tilde{t}_i(\mathbf{x}) d\mathbf{x} \quad (11)$$

This algorithm always has a fixed point, and sometimes has several. If initialized too far away from a fixed point, it may diverge. This is discussed in section 3.3.

### 3.1 THE CLUTTER PROBLEM

For the clutter problem of the previous section, the EP algorithm is

1. The term approximations have the form

$$\tilde{t}_i(\mathbf{x}) = s_i \exp(-\frac{1}{2v_i}(\mathbf{x} - \mathbf{m}_i)^T(\mathbf{x} - \mathbf{m}_i)) \quad (12)$$

   Initialize the prior term to itself: $v_0 = 100$, $\mathbf{m}_0 = 0$, $s_0 = (2\pi v_0)^{-d/2}$. Initialize the data terms so that $\tilde{t}_i(\mathbf{x}) = 1$: $v_i = \infty$, $\mathbf{m}_i = 0$, and $s_i = 1$.

2. $\mathbf{m}_x = \mathbf{m}_0, v_x = v_0$

3. Until all $(\mathbf{m}_i, v_i, s_i)$ converge (changes are less than $10^{-4}$):

   loop $i = 1, ..., n$:

   (a) Remove $\tilde{t}_i$ from the posterior to get an 'old' posterior:

   $$\begin{aligned} (v_x^{\backslash i})^{-1} &= v_x^{-1} - v_i^{-1} \\ \mathbf{m}_x^{\backslash i} &= \mathbf{m}_x + v_x^{\backslash i} v_i^{-1}(\mathbf{m}_x - \mathbf{m}_i) \end{aligned}$$

   (b) Recompute $(\mathbf{m}_x, v_x, Z_i)$ from $(\mathbf{m}_x^{\backslash i}, v_x^{\backslash i})$ as in ADF.
   (c) Update $\tilde{t}_i$:

   $$\begin{aligned} v_i^{-1} &= v_x^{-1} - (v_x^{\backslash i})^{-1} \\ \mathbf{m}_i &= \mathbf{m}_x^{\backslash i} + (v_i + v_x^{\backslash i})(v_x^{\backslash i})^{-1}(\mathbf{m}_x - \mathbf{m}_x^{\backslash i}) \\ s_i &= \frac{Z_i}{(2\pi v_i)^{d/2} \mathcal{N}(\mathbf{m}_i; \mathbf{m}_x^{\backslash i}, (v_i + v_x^{\backslash i})\mathbf{I})} \end{aligned}$$

4. Compute the normalizing constant:

$$B = \frac{\mathbf{m}_x^T \mathbf{m}_x}{v_x} - \sum_i \frac{\mathbf{m}_i^T \mathbf{m}_i}{v_i}$$

$$p(D) \approx (2\pi v_x)^{d/2} \exp(B/2) \prod_{i=0}^{n} s_i$$

Because the term approximations start at 1, the result after one pass through the data is identical to ADF.

## 3.2 RESULTS

EP for the clutter problem is compared with four other algorithms for approximate inference: Laplace's method, variational Bayes, importance sampling (using the prior as the importance distribution), and Gibbs sampling (by introducing hidden variables that determine if a data point is clutter). The goal is to estimate the evidence $p(D)$ and the posterior mean $E[\mathbf{x}|D]$. Figure 1 shows the results on a typical run with $n = 20$ and with $n = 200$. It plots the accuracy vs. cost of the algorithms. Accuracy is measured by absolute difference from the true evidence or the true posterior mean. Cost is measured by the number of floating point operations (FLOPS) in Matlab, via Matlab's `flops` function. This is better than using CPU time because FLOPS ignores interpretation overhead.

The deterministic methods EP, Laplace, and VB all try to approximate the posterior with a Gaussian, so they improve substantially with more data (the posterior is more Gaussian with more data). The sampling methods assume very little about the posterior and cannot exploit the fact that it is becoming more Gaussian. However, this is an advantage for sampling when the posterior has a complex shape. Figure 2 shows an atypical run with a small amount of data ($n = 20$) where the true posterior has three distinct modes. Regular EP did not converge, but a restricted version did (Minka, 2001). Unfortunately, all of the deterministic methods converge to an erroneous result that captures only a single mode.

## 3.3 CONVERGENCE

The EP iterations can be shown to always have a fixed point when the approximations are in an exponential family. The proof is analogous to Yedidia et al. (2000). Let the sufficient statistics be $f_1(\mathbf{x}), ..., f_J(\mathbf{x})$ so that the family has form $\exp(\sum_{j=1}^{J} f_j(\mathbf{x})\lambda_j)$. In the clutter problem we had $f_1(\mathbf{x}) = \mathbf{x}$ and $f_2(\mathbf{x}) = \cdot \mathbf{x}^T\mathbf{x}$. When we treat the prior exactly, the final approximation will be $q(\mathbf{x}) \propto p(\mathbf{x})\exp(\sum_j f_j(\mathbf{x})\nu_j)$ for some $\nu$, and the leave-one-out approximations will be $q^{\backslash i}(\mathbf{x}) \propto p(\mathbf{x})\exp(\sum_j f_j(\mathbf{x})\lambda_{ij})$ for some $\lambda$. Let $n$ be the number of terms $t_i(\mathbf{x})$.

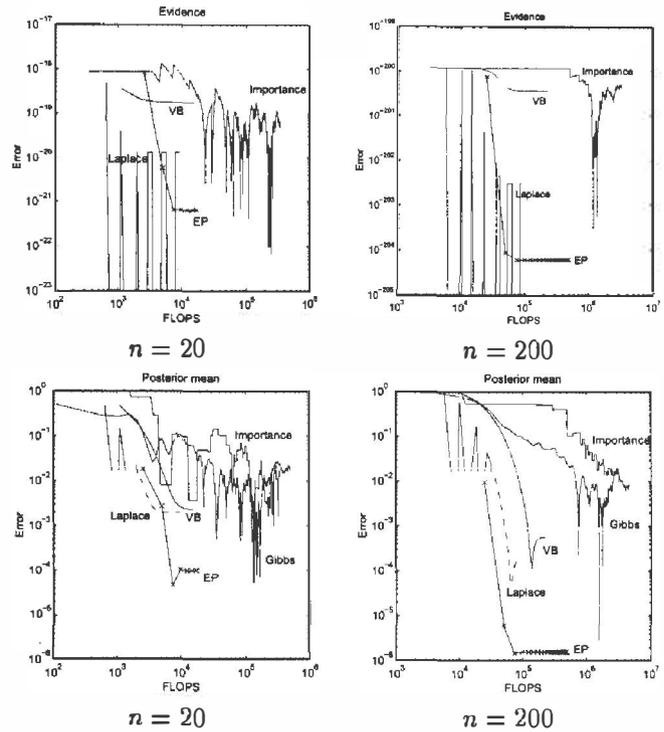

Figure 1: Cost vs. accuracy curves for expectation propagation (EP), Laplace's method, variational Bayes (VB), importance sampling, and Gibbs sampling on the clutter problem with $w = 0.5$ and $\mathbf{x} = 2$. Each 'x' is one iteration of EP. ADF is the first 'x'.

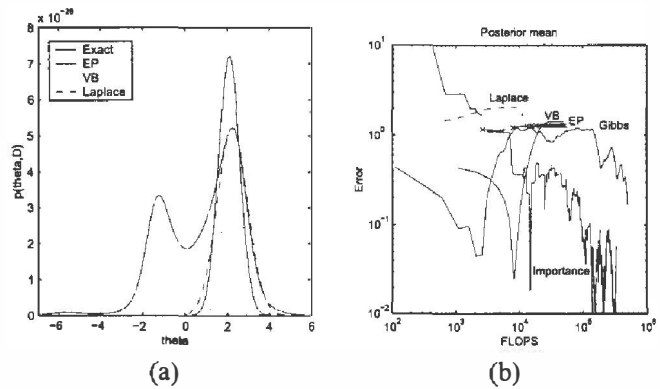

Figure 2: A complex posterior in the clutter problem. (a) Exact posterior vs. approximations obtained by EP, Laplace's method, and variational Bayes. (b) Cost vs. accuracy.

The EP fixed points are in one-to-one correspondence with stationary points of the objective

$$\min_\nu \max_\lambda (n-1)\log\int_\mathbf{x} p(\mathbf{x})\exp(\sum_j f_j(\mathbf{x})\nu_j)d\mathbf{x}$$
$$-\sum_{i=1}^n \log\int_\mathbf{x} t_i(\mathbf{x})p(\mathbf{x})\exp(\sum_j f_j(\mathbf{x})\lambda_{ij})d\mathbf{x} \quad (13)$$
$$\text{such that } (n-1)\nu_j = \sum_i \lambda_{ij} \quad (14)$$

Note that min-max cannot be exchanged with max-min in this objective. By taking derivatives we get the stationary conditions $\int_\mathbf{x} f_j(\mathbf{x})q(\mathbf{x})d\mathbf{x} = \int_\mathbf{x} f_j(\mathbf{x})\hat{p}(\mathbf{x})$, where $\hat{p}(\mathbf{x})$ is defined by (4). This is an EP fixed point. In reverse, given an EP fixed point we can recover $\nu$ and $\lambda$ from $q(\mathbf{x})$ and $q^{\backslash i}(\mathbf{x})$ to obtain a stationary point of (13).

Assume all terms are bounded: $t_i(\mathbf{x}) \leq c$. Then the objective is bounded from below, because for any $\nu$ we can choose $\lambda_{ij} = \frac{n-1}{n}\nu_j$, and then the second part of (13) is at least

$$-n\log\int_\mathbf{x} cp(\mathbf{x})\exp(\sum_j f_j(\mathbf{x})\nu_j)^{\frac{n-1}{n}}d\mathbf{x}$$
$$\geq -n\log c - (n-1)\log\int_\mathbf{x} p(\mathbf{x})\exp(\sum_j f_j(\mathbf{x})\nu_j)d\mathbf{x}$$

by the concavity of the function $y^{\frac{n-1}{n}}$. Therefore there must be stationary points. Sometimes there are multiple fixed points of EP, in which case we can define the 'best' fixed point as the one with minimum energy (13). When canonical EP does not converge, we can minimize (13) by some other scheme, such as gradient descent. In practice, it is found that when canonical EP does not converge, it is for a good reason, namely the approximating family is a poor match to the exact posterior. This happened in the previous example. So before considering alternate ways to carry out EP, one should reconsider the approximating family.

## 4 LOOPY BELIEF PROPAGATION

Expectation Propagation and assumed-density filtering can be used to approximate a belief network by a simpler network with fewer edges. This section shows that if the approximation is completely disconnected, then ADF yields the algorithm of Boyen & Koller (1998) and EP yields loopy belief propagation.

Let the hidden variables be $x_1, ..., x_K$ and collect the observed variables into $\mathbf{D} = \{y_1, ..., y_N\}$. A completely disconnected distribution for $\mathbf{x}$ has the form

$$q(\mathbf{x}) = \prod_{k=1}^K q_k(x_k) \quad (15)$$

When we minimize the KL-divergence $D(\hat{p}(\mathbf{x})||q(\mathbf{x}))$, we will simply preserve the marginals of $\hat{p}(\mathbf{x})$. This corresponds to an expectation constraint $E_q[\delta(x_k - v)] = E_{\hat{p}}[\delta(x_k - v)]$ for all values $v$ of $x_k$. From this we arrive at the ADF algorithm of Boyen & Koller (1998):

1. Initialize $q_k(x_k) = 1$

2. For each term $t_i(\mathbf{x})$ in turn, set $q_k$ to the $k$th marginal of $\hat{p}$:

$$q_k(x_k) = \sum_{\mathbf{x}\backslash x_k} \hat{p}(\mathbf{x}) = \frac{1}{Z_i}\sum_{\mathbf{x}\backslash x_k} t_i(\mathbf{x})q^{\backslash i}(\mathbf{x})$$
$$\text{where } Z_i = \sum_\mathbf{x} t_i(\mathbf{x})q^{\backslash i}(\mathbf{x})$$

For dynamic Bayesian networks, Boyen & Koller set $t_i$ to the product of all of the conditional probability tables for timeslice $i$. Now let's turn this into an EP algorithm. From the ratio $q/q^{\backslash i}$, we see that the approximate terms $\tilde{t}_i(\mathbf{x})$ are completely disconnected. The EP algorithm is thus:

1. $\tilde{t}_i(\mathbf{x}) = \prod_k \tilde{t}_{ik}(x_k)$. Initialize $\tilde{t}_i(\mathbf{x}) = 1$.

2. $q_k(x_k) \propto \prod_i \tilde{t}_{ik}(x_k)$

3. Until all $\tilde{t}_i$ converge:

    (a) Choose a $\tilde{t}_i$ to refine
    (b) Remove $\tilde{t}_i$ from the posterior. For all $k$:

    $$q_k^{\backslash i}(x_k) \propto \frac{q_k(x_k)}{\tilde{t}_{ik}(x_k)} \propto \prod_{j\neq i}\tilde{t}_{jk}(x_k)$$

    (c) Recompute $q(\mathbf{x})$ from $q^{\backslash i}(\mathbf{x})$ as in ADF.
    (d)

    $$\tilde{t}_{ik}(x_k) = Z_i\frac{q_k(x_k)}{q_k^{\backslash i}(x_k)} = \sum_{\mathbf{x}\backslash x_k} t_i(\mathbf{x})\prod_{j\neq k} q_j^{\backslash i}(x_j)$$

To make this equivalent to belief propagation, the original terms $t_i$ should correspond to the conditional probability tables of a directed belief network. That is, we should break the joint distribution $p(D, \mathbf{x})$ into

$$p(D, \mathbf{x}) = \prod_k p(x_k|\mathrm{pa}(x_k)) \prod_j p(y_j|\mathrm{pa}(y_j)) \qquad (16)$$

where $\mathrm{pa}(X)$ is the set of parents of node $X$. The network has observed nodes $y_j$ and hidden nodes $x_k$. The parents of an observed node might be hidden, and vice versa. For an undirected network, the terms are the clique potentials. The quantities in EP now have the following interpretations:

- $q_k(x_k)$ is the belief state of node $x_k$, i.e. the product of all messages into $x_k$.

- The 'old' posterior $q_k^{\setminus i}(x_k)$ for a particular term $i$ is a partial belief state, i.e. the product of messages into $x_k$ except for those originating from term $i$.

- When $i \neq k$, the function $\tilde{t}_{ik}(x_k)$ is the message that node $i$ (either hidden or observed) sends to its parent $x_k$ in belief propagation. For example, suppose node $i$ is hidden and $t_i(\mathbf{x}) = p(x_i|\mathrm{pa}(x_i))$. The other parents send their partial belief states, which the child combines with its partial belief state:

$$\tilde{t}_{ik}(y) = \sum_{\mathbf{x}\setminus x_k} p(x_i|\mathrm{pa}(x_i)) q_i^{\setminus i}(x_i) \prod_{\text{parents } j \neq k} q_j^{\setminus i}(x_j)$$

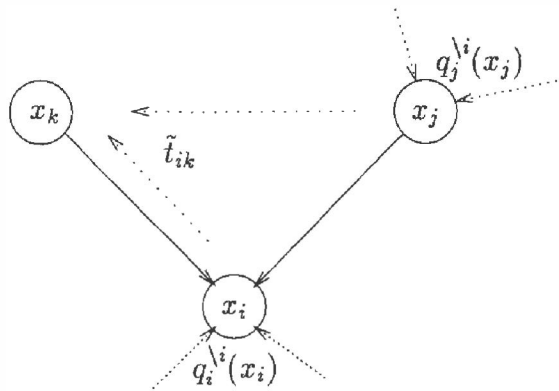

- When node $i$ is hidden, the function $\tilde{t}_{ii}(x_i)$ is a combination of messages sent to node $i$ from its parents in belief propagation. Each parent sends it partial belief state, and the child combines them according to

$$\tilde{t}_{ii}(x_i) = \sum_{\mathrm{pa}(x_i)} p(x_i|\mathrm{pa}(x_i)) \prod_{\text{parents } j} q_j^{\setminus i}(x_j)$$

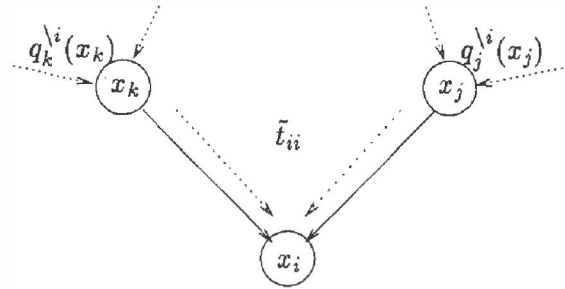

Unlike Pearl's derivation of belief propagation in terms of $\lambda$ and $\pi$ messages, this derivation is symmetric with respect to parents and children. In fact, it is the form used in in factor graphs (Kschischang et al., 2000). All of the nodes that participate in a conditional probability table $p(X|\mathrm{pa}(X))$ send messages to each other based on their partial belief states.

Loopy belief propagation does not always converge, but from section 3.3 we know how we could find a fixed point. For an undirected network with pairwise potentials, the EP energy function (13) is a dual representation of the Bethe free energy given by Yedidia et al. (2000).

Alternatively, we can fit an approximate network which is not completely disconnected, such as a tree-structured network. This was done in the ADF context by Frey et al. (2000). A general algorithm for tree-structured approximation using EP is given by Minka (2001).

## 5 BAYES POINT MACHINE

This section applies Expectation Propagation to inference in the Bayes Point Machine (Herbrich et al., 1999). The Bayes Point Machine (BPM) is a Bayesian approach to linear classification. A linear classifier classifies a point $\mathbf{x}$ according to $y = \mathrm{sign}(\mathbf{w}^T\mathbf{x})$ for some parameter vector $\mathbf{w}$ (the two classes are $y = \pm 1$). Given a training set $D = \{(\mathbf{x}_1, y_1), ..., (\mathbf{x}_n, y_n)\}$, the likelihood for $\mathbf{w}$ can be written

$$p(D|\mathbf{w}) = \prod_i p(y_i|\mathbf{x}_i, \mathbf{w}) = \prod_i \phi\left(\frac{y_i \mathbf{w}^T \mathbf{x}_i}{\epsilon}\right) \qquad (17)$$

$$\phi(z) = \int_{-\infty}^z \mathcal{N}(z; 0, 1) dz \qquad (18)$$

By using $\phi$ instead of a step function, this likelihood tolerates small errors. The allowed 'slack' is controlled by $\epsilon$. To avoid estimating $\epsilon$, which is tangential to this paper, the experiments all use $\epsilon \to 0$, where $\phi$ becomes a step function. The BPM is a hybrid belief network, with $\mathbf{w}$ and $\mathbf{x}_i$ pointing to $\mathbf{y}_i$. Under the Bayesian approach, we also have a prior distribution on $\mathbf{w}$, which is taken to be $\mathcal{N}(\mathbf{0}, \mathbf{I})$. Given this model, the optimal way to classify a new data point $\mathbf{x}$ is to vote all classifiers according to their posterior probability: $E[\text{sign}(\mathbf{w}^T \mathbf{x})|D]$. As an approximation to this, the BPM uses the output of the average classifier: $\text{sign}(E[\mathbf{w}]^T \mathbf{x})$.

Using EP, we can make a multivariate Gaussian approximation to the posterior over $\mathbf{w}$ and use its mean as the estimated Bayes point. The resulting algorithm is similar to that for the clutter problem. To save notation, $y_i \mathbf{x}_i / \epsilon$ is written simply as $\mathbf{x}_i$.

1. $\tilde{t}_i(\mathbf{w}) = s_i \exp(-\frac{1}{2v_i}(\mathbf{w}^T \mathbf{x}_i - m_i)^2)$. Initialize with $v_i = \infty, m_i = 0, s_i = 1$.

2. $q(\mathbf{w}) = \mathcal{N}(\mathbf{m}_w, \mathbf{V}_w)$. Initialize with the prior: $\mathbf{m}_w = \mathbf{0}, \mathbf{V}_w = \mathbf{I}$.

3. Until all $(m_i, v_i)$ converge (changes are less than $10^{-4}$):

    loop $i = 1, ..., n$:

    (a) Remove $\tilde{t}_i$ from the posterior to get an 'old' posterior:
    $$\mathbf{V}_w^{\setminus i} = \mathbf{V}_w + \frac{(\mathbf{V}_w \mathbf{x}_i)(\mathbf{V}_w \mathbf{x}_i)^T}{v_i - \mathbf{x}_i^T \mathbf{V}_w \mathbf{x}_i}$$
    $$\mathbf{m}_w^{\setminus i} = \mathbf{m}_w + (\mathbf{V}_w^{\setminus i} \mathbf{x}_i) v_i^{-1} (\mathbf{x}_i^T \mathbf{m}_w - m_i)$$

    (b) Recompute $(\mathbf{m}_w, \mathbf{V}_w)$ from $(\mathbf{m}_w^{\setminus i}, \mathbf{V}_w^{\setminus i})$, using ADF:
    $$z_i = \frac{(\mathbf{m}_w^{\setminus i})^T \mathbf{x}_i}{\sqrt{\mathbf{x}_i^T \mathbf{V}_w^{\setminus i} \mathbf{x}_i + 1}}$$
    $$\alpha_i = \frac{1}{\sqrt{\mathbf{x}_i^T \mathbf{V}_w^{\setminus i} \mathbf{x}_i + 1}} \frac{\mathcal{N}(z_i; 0, 1)}{\phi(z_i)}$$
    $$\mathbf{m}_w = \mathbf{m}_w^{\setminus i} + \mathbf{V}_w^{\setminus i} \alpha_i \mathbf{x}_i$$
    $$\mathbf{V}_w = \mathbf{V}_w^{\setminus i} - (\mathbf{V}_w^{\setminus i} \mathbf{x}_i) \left( \frac{\alpha_i \mathbf{x}_i^T \mathbf{m}_w}{\mathbf{x}_i^T \mathbf{V}_w^{\setminus i} \mathbf{x}_i} \right) (\mathbf{V}_w^{\setminus i} \mathbf{x}_i)^T$$

    (c) Update $\tilde{t}_i$:
    $$v_i = \mathbf{x}_i^T \mathbf{V}_w^{\setminus i} \mathbf{x}_i \left( \frac{1}{\alpha_i \mathbf{x}_i^T \mathbf{m}_w} - 1 \right)$$
    $$m_i = \mathbf{x}_i^T \mathbf{m}_w^{\setminus i} + (v_i + \mathbf{x}_i^T \mathbf{V}_w^{\setminus i} \mathbf{x}_i) \alpha_i$$

$$s_i = \frac{\phi(z_i) \sqrt{1 + v_i^{-1} \mathbf{x}_i^T \mathbf{V}_w^{\setminus i} \mathbf{x}_i}}{\exp(-\frac{1}{2} \frac{\mathbf{x}_i^T \mathbf{V}_w^{\setminus i} \mathbf{x}_i}{\mathbf{x}_i^T \mathbf{m}_w} \alpha_i)}$$

4. $B = \mathbf{m}_w^T \mathbf{V}_w^{-1} \mathbf{m}_w - \sum_i \frac{m_i^2}{v_i}$
   $p(D) \approx |\mathbf{V}_w|^{1/2} \exp(B/2) \prod_{i=1}^n s_i$

This algorithm processes each data point in $O(d^2)$ time. Assuming the number of iterations is constant, which seems to be true in practice, computing the Bayes point therefore takes $O(nd^2)$ time. This algorithm can be extended to use an arbitrary inner product function, just as in Gaussian process classifiers and the Support Vector Machine, which changes the running time to $O(n^3)$, regardless of dimensionality. This extension can be found in Minka (2001). Interestingly, Opper & Winther (2000) have derived an equivalent algorithm using statistical physics methods. However, the EP updates tend to be faster than theirs and do not require a stepsize parameter.

### 5.1 RESULTS

Figure 3(a) demonstrates the Bayes point classifier vs. the SVM classifier on 3 training points. Besides the two dimensions shown here, each point had a third dimension set at 1. This provides a 'bias' coefficient $w_3$ so that the decision boundary doesn't have to pass through $(0, 0)$. The Bayes point classifier approximates a vote between all linear separators, ranging from an angle of 0° to 135°. The Bayes point is an angle in the middle of this range.

Figure 3(b) plots cost vs. error for EP versus three other algorithms for estimating the Bayes point: the billiard algorithm of Herbrich et al. (1999), the TAP algorithm of Opper & Winther (2000), and the mean-field (MF) algorithm of Opper & Winther (2000). The error is measured by Euclidean distance to the exact solution found by importance sampling. The error in using the SVM solution is also plotted for reference. Its unusually long running time is due to Matlab's `quadprog` solver. TAP and MF were slower to converge than EP, even with a large initial step size of 0.5. As expected, EP and TAP converge to the same solution.

Figure 4 compares the error rate of EP, Billiard, and SVM on four datasets from the UCI repository (Blake & Merz, 1998). Each dataset was randomly split 40 times into a training set and test set, in the ratio 60%:40%. In each trial, the features were normalized to have zero mean and unit variance in the training set. The classifiers used zero slack and a Gaussian inner product with standard deviation 3. Billiard was run for 500 iterations. The `thyroid` dataset was made into a binary classification problem by merging the different classes into normal vs. abnormal. Except for `sonar`, EP has lower average error than the SVM (with 99% probability), and in all cases EP is at least as good as Billiard. Billiard has the highest running time because it is initialized at the SVM solution.

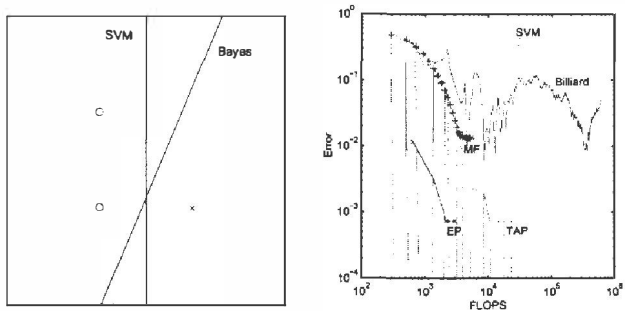

Figure 3: (left) Bayes point machine vs. Support Vector Machine on a simple data set. The Bayes point more closely approximates a vote between all linear separators of the data. (right) Cost vs. error in estimating the posterior mean. ADF is the first 'x' on the EP curve.

| Dataset | EP | Billiard | SVM |
| --- | --- | --- | --- |
| Heart | .203 ± .069 | .207 ± .069 | .232 ± .069 |
| Thyroid | .037 ± .037 | .037 ± .038 | .053 ± .035 |
| Ionosphere | .099 ± .057 | .113 ± .064 | .115 ± .066 |
| Sonar | .140 ± .077 | .147 ± .072 | .129 ± .075 |

Figure 4: Test error rate for the Bayes Point Machine (using EP or the Billiard algorithm) compared to the Support Vector Machine. Reported is the mean over 40 train-test splits ± two standard deviations. These results are for a Gaussian kernel with $\sigma = 3$, and will differ with other kernels.

## 6 SUMMARY

This paper presented a generalization of belief propagation which is appropriate for hybrid belief networks. Its superior speed and accuracy were demonstrated on a Gaussian mixture network and the Bayes Point Machine. Hopefully it will prove useful for other networks as well. The Expectation Propagation iterations always have a fixed point, which can be found by minimizing an energy function.

## Acknowledgment

This work was performed at the MIT Media Lab, supported by the Things That Think and Digital Life consortia.